# DEVNAGARI DOCUMENT SEGMENTATION USING HISTOGRAM APPROACH

Vikas J Dongre [1]  Vijay H Mankar [2]

Department of Electronics & Telecommunication,
Government Polytechnic, Nagpur, India
[1]dongrevj@yahoo.co.in; [2]vhmankar@gmail.com

## ABSTRACT

*Document segmentation is one of the critical phases in machine recognition of any language. Correct segmentation of individual symbols decides the accuracy of character recognition technique. It is used to decompose image of a sequence of characters into sub images of individual symbols by segmenting lines and words. Devnagari is the most popular script in India. It is used for writing Hindi, Marathi, Sanskrit and Nepali languages. Moreover, Hindi is the third most popular language in the world. Devnagari documents consist of vowels, consonants and various modifiers. Hence proper segmentation of Devnagari word is challenging. A simple histogram based approach to segment Devnagari documents is proposed in this paper. Various challenges in segmentation of Devnagari script are also discussed.*

## 1. INTRODUCTION

Machine learning and human computer interaction are the most challenging research fields since the evolution of digital computers. In Optical Character Recognition (OCR), the text lines, words and symbols in a document must be segmented properly before recognition. Correctness/ incorrectness of text line segmentation directly affect accuracy of word/character segmentation and consequently affect the accuracy of word/character recognition [1]. Several techniques for text line segmentation are reported in the literature [2-6]. These techniques may be classified into three groups as follows: (i) Projection profile based techniques, (ii) Hough transform based techniques, (iii) Thinning based approach. As a conventional technique for text line segmentation, global horizontal projection analysis of black pixels has been utilized in [4, 7]. Piece-wise horizontal projection analysis of black pixels is employed by many researchers to segment text pages of different languages [2, 9]. In piecewise horizontal projection technique, the text-page image is decomposed into horizontal stripes. The positions of potential piece-wise separating lines are obtained for each stripe using horizontal projection on each stripe. The potential separating lines are then connected to achieve complete separating lines for all respective text lines located in the text page image. Concept of the Hough transform is employed in the field of document analysis in many research areas such as skew detection, slant detection, text line segmentation, etc [8]. Thinning operation is also used by researchers for text line segmentation from documents [10].

In this paper we have proposed a bounded box method for segmentation of documents lines and words and characters. The method is based on the pixel histogram obtained. The organization of this paper is as follows: In Section 2, we have discussed features of Indian scripts. Section 3 discusses image preprocessing methods. Section 4 details the proposed segmentation approach. Experimental results are discussed in Section 5 and scope for further research is discussed in Section 6.

## 2. FEATURES OF DEVNAGARI SCRIPT

India is a multi-lingual and multi-script country comprising of eighteen official languages. Because there is typically a letter for each of the phonemes in Indian languages, the alphabet set tends to be quite large. Hindi, the national language of India, is written in the Devnagari script. Devnagari is also used for writing Marathi, Sanskrit and Nepali. Moreover, Hindi is the third most popular language in the world [1]. It is spoken by more than 500 million people in the world. Devnagari has 11 vowels and 33 consonants. They are called basic characters. Vowels can be written as independent letters, or by using a variety of diacritical marks which are written above, below, before or after the consonant they belong to. When vowels are written in this way they are known as modifiers and the characters so formed are called conjuncts. Sometimes two or more consonants can combine and take new shapes. These new shaped clusters are known as compound characters. These types of basic characters, compound characters and modifiers are present not only in Devnagari but also in other scripts.

All the characters have a horizontal line at the upper part, known as *Shirorekha*. In continuous handwriting, from left to right direction, the shirorekha of one character joins with the shirorekha of the previous or next character of the same word. In this fashion, multiple characters and modified shapes in a word appear as a single connected component joined through the common shirorekha. Also in Devnagari there are vowels, consonants, vowel modifiers and compound characters, numerals. Moreover, there are many similar shaped characters. All these variations make Devnagari Optical Character Recognition, a challenging problem. A sample of Devnagari character set is provided in table 1 to 6.

Table 1: Vowels and Corresponding Modifiers

| Vowels: | अ | आ | इ | ई | उ | ऊ | ऋ | ए | ऐ | ओ | औ |
|---|---|---|---|---|---|---|---|---|---|---|---|
| Modifiers: | | ा | ि | ी | ु | ू | ृ | े | ै | ो | ौ |

Table 2: Consonants

| क | ख | ग | घ | ङ | च | छ | ज | झ | ञ | ट |
|---|---|---|---|---|---|---|---|---|---|---|
| ठ | ड | ढ | ण | त | थ | द | ध | न | प | फ |
| ब | भ | म | य | र | ल | व | श | ष | स | ह |

Table 3: Half Form of Consonants with Vertical Bar.

| क् | ख् | ग् | | घ् | | ज् | झ् | ञ् |
|---|---|---|---|---|---|---|---|---|
| | | ण् | त् | थ् | | ध् | न् | प् | फ् |
| ब् | भ् | म् | य् | | ल् | व् | श् | ष् | स् |

Table 4: Examples of Combination of Half-Consonant and Consonant.

| क्क | क्स | क्ल | | क्त | घ्न | ध्व | ग्य | घ्य | ग्ज्ञ | त्न | त्न | प्त | प्न | प्ल | ल्न |
|---|---|---|---|---|---|---|---|---|---|---|---|---|---|---|---|
| द्व | द्ब | | ध्न | ध्म | | म्ल | म्न | | ल्ल | ल्म | | श्न | श्म | | श्व | श्च | | श्ल | श्म | | ष्ण | ष्म |

Table 5: Examples of Special Combination of Half-Consonant and Consonant.

| क्ष | प्क्ष | | ज्ञ | ज्ज | | ट्ट | ट्ठ | | ट्ठ | ट्ठ | | त्र | ल्त | | द्द | द्ड |
|---|---|---|---|---|---|---|---|---|---|---|---|---|---|---|---|
| द्ध | ध्ब्द | | द्व | द्द्र | | द्व | द्र्ब्द | | श्र | श्र | | द्भ | ब्ज | | द्य | द्ग्य |

Table 6: Special Symbols

| क | ख | ग | ज | फ | इ | ढ़ | ँ | : | । | ऽ | ॱ |
|---|---|---|---|---|---|---|---|---|---|---|---|

## 3. IMAGE PREPROCESSING

We have collected the printed pages from different office correspondence. The document pages are scanned using a flat bed scanner at a resolution of 300 dpi. These pixels may have values: OFF (0) or ON (1) for binary images, 0–255 for gray-scale images, and 3 channels of 0–255 colour values for colour images. Colour image is converted to grayscale by eliminating the hue and saturation information while retaining the luminance. It is further analyzed to get useful information. Such processing is explained below.

### 3.1 Thresholding and Binarization:

The digitized text images are converted into binary images by thresholding using Otsu's method [17]. Original image contains 0 for Object and 1 for background. The image inverted to obtain image such that object pixels are represented by 1 and background pixels by 0.

### 3.2 Noise reduction:

The noise, introduced by the optical scanning device or the writing instrument, causes disconnected line segments, bumps and gaps in lines, filled loops etc. The distortion including local variations, rounding of corners, dilation and erosion is also a problem. Prior to the character recognition, it is necessary to eliminate these imperfections [11-12]. It is carried using various morphological processing techniques.

### 3.3 Skew Detection and Correction:

Handwritten document may originally be skewed or skewness may introduce in document scanning process. This effect is unintentional in many real cases, and it should be eliminated because it dramatically reduces the accuracy of the subsequent processes such as segmentation and classification. Skewed lines are made horizontal by calculating skew angle and making proper correction in the raw image using Hu moments and various transforms [13-15].

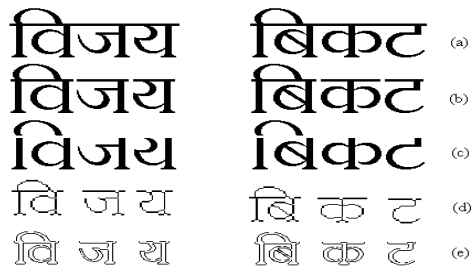

Figure 1: Preprocessed Images (a) Original, (b) segmented (c) Shirorekha removed (d) Thinned (e) image edging

### 3.4 Thinning:

The boundary detection of image is done to enable easier subsequent detection of pertinent features and objects of interest (see fig.1- a to e). Various standard functions are available in MATLAB for above operations [16].

# 4. PROPOSED SEGMENTATION APPROACH

After the image is preprocessed using methods discussed in section 3, we now apply various techniques for segmentation of document lines, words and characters. The process of segmentation mainly follows the following pattern:

1) Identify the text lines in the page.
2) Identify the words in individual line.
3) Finally identify individual character in each word.

## 4.1 Line Segmentation.

The global horizontal projection method is used to compute sum of all white pixels on every row and construct corresponding histogram. The steps for line segmentation are as follow:

- Construct the Horizontal Histogram for the image (fig. 2-b).
- Count the white pixel in each row.
- Using the Histogram, find the rows containing no white pixel.
- Replace all such rows by 1 (fig. 2-c).
- Invert the image to make empty rows as 0 and text lines will have original pixels.
- Mark the Bounding Box for text lines (figure 2-e) using standard Matlab functions (regionprops and rectangle).
- Copy the pixels in Bounding Box and save in separate file. (Separated lines shown in fig. 2-f).

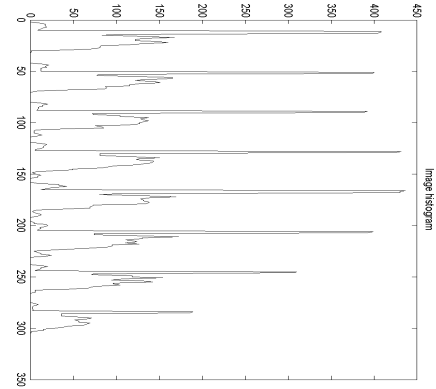

<div>
a) Original Scanned Document        (b) Image Histogram
</div>

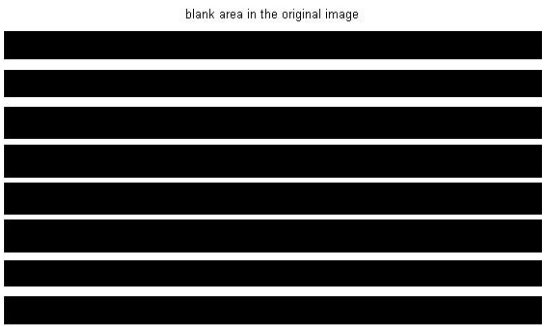
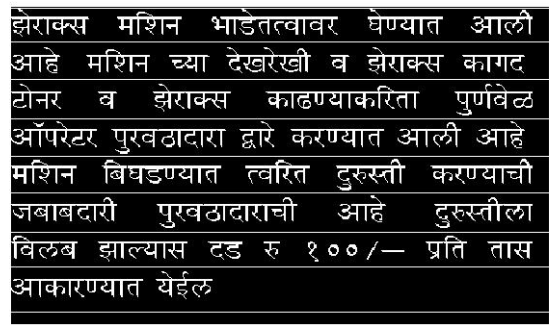

<div>
(c) Blank space between the lines        (d) Line separation
</div>

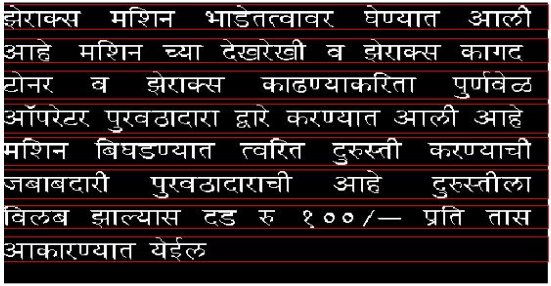 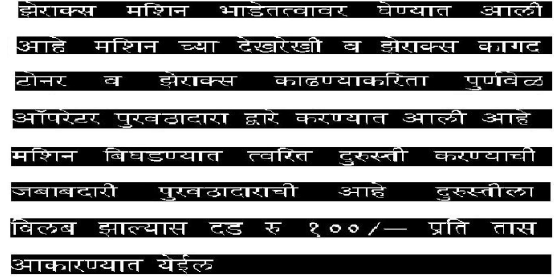

(e) Regions of interest                          (f) segmented lines

Figure 2:   Line Segmentation

## 4.2 Word Segmentation

The global horizontal projection method is used here to compute sum of all white pixels on every column and construct corresponding histogram. The steps for line segmentation are as follow:

- Construct the Vertical Histogram for the image (fig. 3-b).
- Count the white pixel in each column.
- Using the Histogram, find the columns containing no white pixel.
- Replace all such columns by 1
- Invert the image to make empty rows as 0 and text words will have original pixels.
- Mark the Bounding Box for word. (See fig 3-c)
- Copy the pixels in the Bounding Box and save in separate file. (See fig. 3-d).

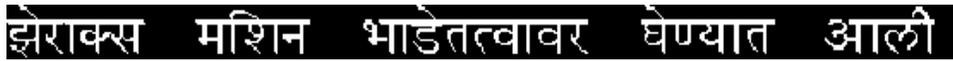

(a) Original line

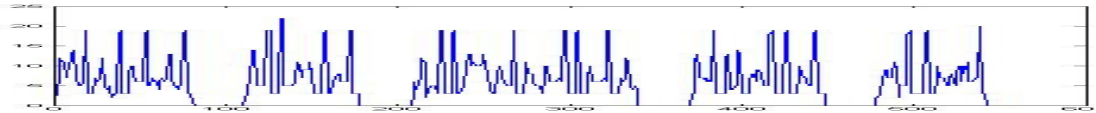

(b) Word Histogram

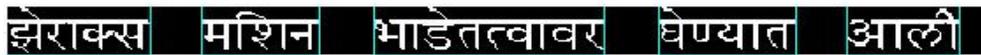

(c)   Regions of interest

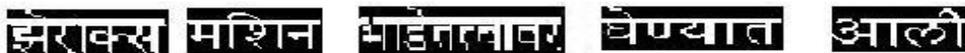

(d) Segmented words

Figure 3:  Word Segmentation

50

## 4.3 Character Segmentation

A slight modification in previous algorithm (section 4.2) is used here. The steps for line segmentation are as follow:

- Get the thinned image using Matlab bwmorph function. (This is done to normalize image against thickness of the character).
- Count the white pixel in each column.
- Find the position containing single white pixel.
- Replace all such columns by 1.
- Invert the image to make such columns as 0 and text characters will have original pixels.
- Mark the Bounding Box for characters using standard Matlab functions. See fig 4-a.
- Copy the pixels in the Bounding Box and save in separate file. (Separated characters are shown in fig. 4-b).

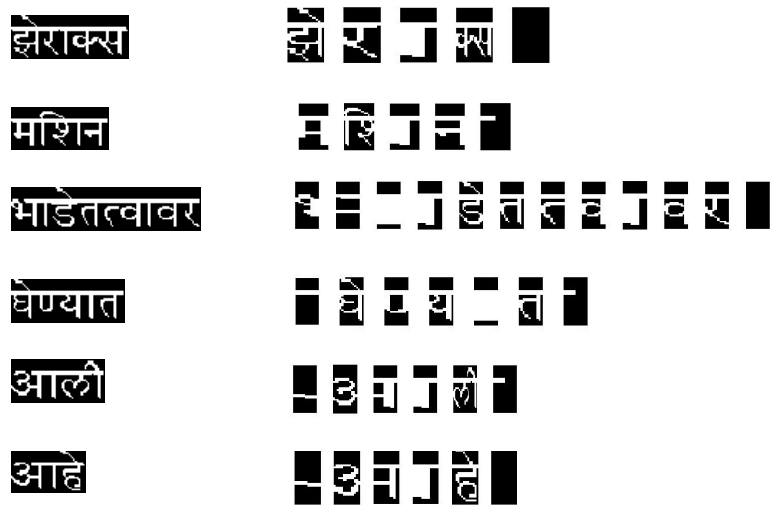

(a) Region of Interest      (b) segmented characters

Figure 4: Character segmentation

## 5. RESULTS AND DISCUSSION

Various documents were collected and tested. It is observed that line segmentation is done with nearly 100% accuracy. Word segmentation is accurate as long as the document contains characters only. When Devnagari numerals are present in the document, which does not contain *shirorekha*, each digit is considered as separate word by the algorithm. Hence accuracy is reduced marginally. In the present case it is 91%.

Table 7: Character Segmentation results for document in fig 2 (a)

| Words ( in figure 4) | 1 | 2 | 3 | 4 | 5 | 6 |
|---|---|---|---|---|---|---|
| Characters present | 3 | 3 | 6 | 3 | 2 | 2 |
| Characters recognized | 5 | 5 | 12 | 7 | 6 | 6 |
| Accuracy | 60 % | 60 % | 50 % | 42 % | 33 % | 33 % |

Table 8:  Overall Segmentation results for document in fig 2 (a)

| Line Segmentation | Lines in Document | Recognized lines | Accuracy |
|---|---|---|---|
| | 8 | 8 | 100 % |
| Word Segmentation | words in Document | Recognized words | Accuracy |
| | 41 | 45 | 91 % |
| Line Segmentation | Characters in Document | Recognized Characters | Accuracy |
| | 133 | 242 | 55 % |

In case of character segmentation, words are segmented into more symbols than actually present in the word as shown in figure 4. Result is summarized in Table 8. This error is resulted since the words are scanned only from top to bottom by the algorithm used. Devnagari is two dimensional script as consonants are modified in many ways from top, bottom, left or right to form a meaningful letter. Unconnected Vertical lines in the words are recognized as separate symbol by the algorithm used. For accurate segmentation, all the modifiers must be segmented so that their recognition can be properly done done.

# 6. CONCLUSIONS AND FUTURE WORK

In this paper, we have presented a primary work for segmentation of lines, words and characters of Devnagari script. Nearly 100% successful segmentation achieved in line and word segmentation but character level segmentation needs more effort as it is complicated for Devnagari script. This is challenging work due to following reasons.

• Compound letters are connected at various places. It is difficult to identify exact connecting points for segmentation.
• Upper and lower modifier segmentation needs different approaches.
• Separating *anuswara* (.) and *full stop* (.) from noise is critical as both resemble the same. Knowledge of natural language processing techniques needs to be applied here.
• Handwritten unconnected compound letter segmentation is also critical.
• Handwritten unintentionally connected simple letter segmentation is also critical.

All these issues will be dealt in the future for printed and handwritten documents in Devnagari script by using various approaches.

## Authors

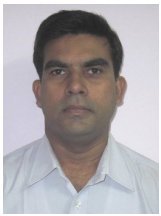

**Vikas J Dongre** received B.E and M.E. in Electronics in 19991 and 1994 respectively. He served as lecturer in SSVPS engineering college Dhule, (M.S.) India from 1992 to 1994. He Joined Government Polytechnic Nagpur as Lecturer in 1994 where he is presently working as lecturer (selection grade). His areas of interests include Microcontrollers, embedded systems, image recognition, and innovative Laboratory practices. He is pursuing for PhD in Offline Handwritten Devnagari Character Recognition. He has published one research paper in international journal and two research paper in international conferences.

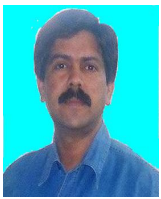

**Vijay H. Mankar** received M. Tech. degree in Electronics Engineering from VNIT, Nagpur University, India in 1995 and Ph.D. (Engg) from Jadavpur University, Kolkata, India in 2009 respectively. He has more than 17 years of teaching experience and presently working as a Lecturer (Selection Grade) in Government Polytechnic, Nagpur (MS), India. He has published more than 30 research papers in international conference and journals. His field of interest includes digital image processing, data hiding and watermarking.